# Multi-Agents Dynamic Case Based Reasoning & Inverse Longest Common Sub-Sequence And Individualized Follow-up of Learners in CEHL

**Abdelhamid Zouhair[1,2], El Mokhtar En-Naimi[1], Benaissa Amami[1], Hadhoum Boukachour[2], Patrick Person[2], Cyrille Bertelle[2]**

**[1] LIST (Laboratoire d'Informatique, Systèmes et Télécommunications)**
**The University of Abdelmalek Essaâdi, FST of Tangier, Tangier, Morocco**

**[2] LITIS (Laboratoire d'Informatique, de Traitement de l'Information et des Systèmes)**
**The University of Le Havre, 25, rue Ph. Lebon, 76600, Le Havre, France**

**Abstract**

In E-learning, there is still the problem of knowing how to ensure an individualized and continuous learner's follow-up during learning process, indeed among the numerous tools proposed, very few systems concentrate on a real time learner's follow-up. Our work in this field develops the design and implementation of a Multi-Agents System Based on Dynamic Case Based Reasoning which can initiate learning and provide an individualized follow-up of learner. When interacting with the platform, every learner leaves his/her traces in the machine. These traces are stored in a basis under the form of scenarios which enrich collective past experience. The system monitors, compares and analyses these traces to keep a constant intelligent watch and therefore detect difficulties hindering progress and/or avoid possible dropping out. The system can support any learning subject. The success of a case-based reasoning system depends critically on the performance of the retrieval step used and, more specifically, on similarity measure used to retrieve scenarios that are similar to the course of the learner (traces in progress). We propose a complementary similarity measure, named Inverse Longest Common Sub-Sequence (ILCSS). To help and guide the learner, the system is equipped with combined virtual and human tutors.

***Keywords***: *Computer Environment for Human Learning, Dynamic Case-Based Reasoning, Multi-Agent Systems, similarity measure, Inverse Longest Common Sub-Sequence (ILCSS), Traces.*

## 1. Introduction

E-learning or Computing Environment for Human Learning (CEHL) is a computer tool which offers learners another medium of learning. Indeed it allows learner to break free from the constraints of time and place of training. They are due to the learner's availability. In addition, the instructor is not physically present and training usually happens asynchronously. However, most E-learning platforms allow the transfer of knowledge in digital format, without integrating the latest teaching approach in the field of education (e. g. constructivism, [26], ...). Consequently, in most cases distance learning systems degenerate into tools for downloading courses in different formats (pdf, word ...) or into sending homework to teachers on servers. These platforms also cause significant overload and cognitive disorientation for learners. Today, it is therefore necessary to design a CEHL that provides individualized follow-up to meet the pace and process of learning for the learner, who thus becomes the pilot of training. The system will also respond to the learner's specific needs. Our contribution in this field is to design and implement a computer system (i. e. intelligent tutor) able to initiate the learning and provide an individualized monitoring of the learner.

Solving these problems involves first, to understand the behavior of the learner, or group of learners, who use CEHL to identify the causes of problems or difficulties which a learner can encounter. This can be accomplished while leaning on the traces of interactions of the learner with the CEHL, which include history, chronology of interactions and productions left by the learner during his/her learning process. This will allow us the reconstruction of perception elements of the activity performed by the learner. According to Marty and Mille [20] the digital traces of interactions represent a major resource customization CEHL. The same authors also add that the theory, the practice protocols development, generic tools, etc., can significantly alter the supply of human learning activities mediated by a computing environment for human learning.

The traces are generally numerous, coming from different sources and with different levels of granularity. Therefore, the observation process-based traces suggest both the collection of traces together with their structure [29].

We propose a system (i. e. intelligent tutor) able to represent, follow and analyze the evolution of a learning situation through the exploitation and the treatment of the traces left by the learner during his/her learning on the platform. This system is based, firstly on the traces to feed the system and secondly on the reconciliation between the course of the learner (traces in progress) and past courses (or past traces). The past traces are stored in the form of scenarios in a database called “base of scenarios”. The analysis of the course must be executed continuously and in real time which leads us to choose a Multi-Agent architecture allowing the implementation of a dynamic case-based reasoning.

Recently, several research works have been focused on the dynamic case based reasoning in order to push the limits of case based reasoning system dealing with situations known as “static”, reactive and responsive to users. All these works are based on the observation that the current tools are limited in capabilities, and are not capable of evolving to fit the non-anticipated or emerging needs. For example, few CBR systems are able to change over time the way of representing a case [7]. According Alain Mille, a case has to describe its context of use, which is very difficult to decide before any reuse and can change in time [22].

The success of a case-based reasoning system depends primarily on the performance of the retrieval step used and, more particularly, on similarity measure used to retrieve scenarios that are similar to the course of the learner (traces in progress). Several research works have been focused on the similarity measure. Furthermore, these methods are not well suited when we compare two heterogeneous sequences containing textual data (we need semantic distance). In addition we must begin to compare the sequences from tail.

In order to deal with this issue, we propose a complementary similarity measure entitled Inverse Longest Common Sub-Sequence an extension of the Longest Common Sub-Sequence measure.

We propose a system, which analyzes the traces of learners in a continuous way, in order to ensure an automatic and a continuous monitoring of the learner. Our work in this field develops the design and implementation of a Dynamic Case Based Reasoning founded on the Multi-Agent Systems.

Several questions arise: How to ensure an individualized and continuous learner’s follow-up during the learning process? How to represent the current situation using the traces of the interaction and how to define its structure? How to implement the case-based reasoning in our situation? Other problems that are related to our choice of case based reasoning approach also arise, such as how to define the structure of cases, case base and the case based reasoning cycle? Finally, we must analyze how to implement the reasoning process of our particular dynamic situation.

The rest of this paper is organized as follows: In the second section, we give a general introduction of E-learning and intelligent tutoring. The third section is devoted to the presentation of the design and implementation of our approach. We will give an overview of the analysis and decomposition needs. So we will introduce the general architecture of the system and we will propose the description of the Multi-Agent Systems (MAS) and its objectives, together with the architecture of our MAS which can implement the approach of dynamic case-based reasoning (DCBR). In section four, we will describe the approach of Case-Based Reasoning and Multi-Agent Case Based Reasoning, in the following part, we will propose the description of our approach in Case Based Reasoning field: Multi-Agent Dynamic Case Based Reasoning and we will propose the description of our contribution in similarity measure entitled Inverse Longest Common SubSequence. In section five we discuss the traces which are left by the learner and feed our system. In addition, we describe the ontology of the learner’s course, semantic features and the proximity measure in order to structure the learner’s activities. Finally, we will give the conclusion and perspectives of this work.

## 2. Intelligent Tutor and Distance Learning

Intelligent Tutoring Systems (ITS) are computer systems designed to assist and facilitate the task of learning for the learner. They have expertise in so far as they know the subject matter taught (domain knowledge), how to teach (pedagogical knowledge) and also how to acquire information on the learner (learner representative).

There is much research concerned with the design and implementation of computer systems to assist a learner in learning. There are, for example, tutors or teaching agents who accompany learners by proposing remedial activities [11]. There are also the agents of support to the group collaboration in the learning [8] encouraging, the learners’ participation and facilitating discussion between them. Other solutions are based on agents that incorporate and seek to make cooperation among various Intelligent Tutoring Systems [6]. The Baghera platform [33], which is a “distance” CEHL exploits the concepts and methods of Multi-Agent approach. Baghera assists learners in their work solving exercise in geometry. They can interact with other learners or teachers (tutors). The teachers can know the progress status of the learner’s work in order to intervene if necessary. These tools of distance learning do not allow an individualized, continuous and real-time learner’s follow-up. They adopt a traditional pedagogical approach (behaviorist) instead of integrating the latest teaching approaches (constructivism and social constructivism [24], [32]). Finally, given the large number of learners who leave their training, the adaptation of learning according to the learner's profile has become indispensable today.
Our contribution consists in proposing an adaptive system to ensure an automatic and a continuous monitoring of the learner. This monitoring is based on cases (dropping out, difficulties met, etc.) past and similar. Moreover, the system is open, scalable and generic to support any learning subject.

# 3. Our Approach: Design and Implementation

## 3.1 Introduction: Analysis and Decomposition Needs

We reconcile analysis of the traces left by the learner's activity in e-learning, and the decision support systems, able to represent, follow in real-time and analyze the evolution of a dynamic situation. Such a system must:

- Represent the current situation;
- Take into account the dynamic changes of the current situation;
- Predict the possible evolution of this situation;
- React according to particular situations (which depend on the learner's profile).

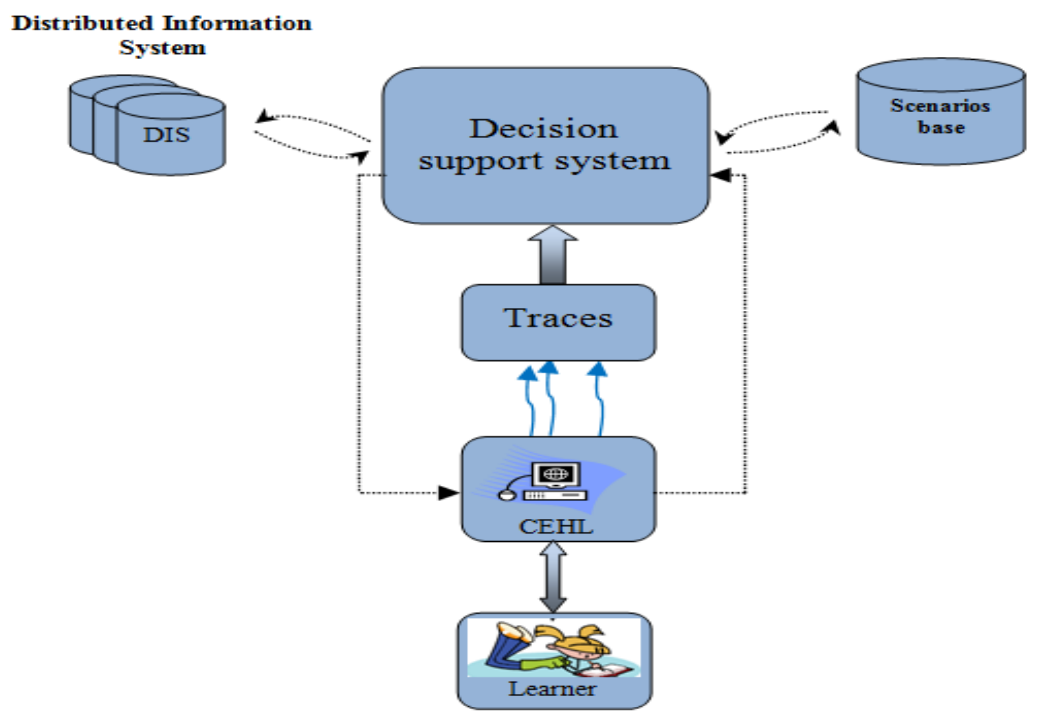

Fig. 1 Decision Support System Architecture.

This can be realized with a study of the traces (past experiences) left by the learner in interaction with the learning platform. Nonetheless, for those past situations we know the consequences that are stored in the memory of our system. This leads us to choose a tool for the formalization of the experience: case-based reasoning (CBR) [17]. In fact, CBR is an approach of artificial intelligence, considered as the most privileged method modeling users' past experience and incremental learning from this experience.

One of the goals of the learner's follow-up individualized is to predict and reduce the number of dropping out, which leads us to seek a flexible and adaptive solution. Such a solution, a decision support system (Figure 1) allows to analyze the course of the learner in order to anticipate a possible dropping-out of the learner or the learning difficulties of the latter. But such a system must take into account:

- The complexity of the situations to be treated;
- The dynamic representation of the current situation;
- The representation of past situations (scenarios);
- The link between current situation (current situation analysis) and scenarios (previous situations).

We propose a system, which analyzes the traces of learners in a continuous way. Moreover, the system must take into account the evolving and dynamic character of the course to be analyzed. The analysis is based on the link, between the course of learner (traces in progress) and the past courses (traces). The traces of past learning activities will be the source of knowledge for the learning adaptation process. They are stored in a database called "base of scenarios". Each scenario contains all the key aspects of its development, that is to say the facts that have played an effective role in how events are unfolded.

## 3.2 General Architecture of the System

**Description of the System and its Objectives**: One of the main objectives of the individualized monitoring of the learner is to envisage, to anticipate and to reduce the number of dropping out, which makes us seek a flexible and adaptive solution [10]. The complexity of the situations to be treated leads us to choose an approach based on a Multi-Agent Systems (MAS), able to cooperate and coordinate their actions to provide a pedagogical adaptation for the learner's profile. We reconcile the problems of the analysis of the traces left by the learner's activity in e-learning, and the decision support systems, able to represent, follow in real-time and analyze the evolution of a dynamic situation. Such a system must represent the current situation, take into account the dynamic change of the current situation, predict the possible evolution of this situation, and react depending on the particular situations and also depend on the learners' profiles. This can be done by using past situations which consequences are known. It is then a question of reasoning by analogy. This type of reasoning can allow solving new problems, using already solved problems available in memory. We often resort to our experience to solve new problems.

The system we propose, allows to analyze the learner's course (trace) in order to anticipate a possible dropping-out. The learning activities past traces will be the source of knowledge for the learning adaptation process, they are stored in a database called ‘’base of scenarios’’. Each scenario contains all determining aspects in its development, i.e, the facts that have played an effective role in the way the events proceeded. The analysis of the current situation must be continuous and dynamic. Indeed, the target case is a plot that evolves, therefore the system must take this incremental evolution into account.

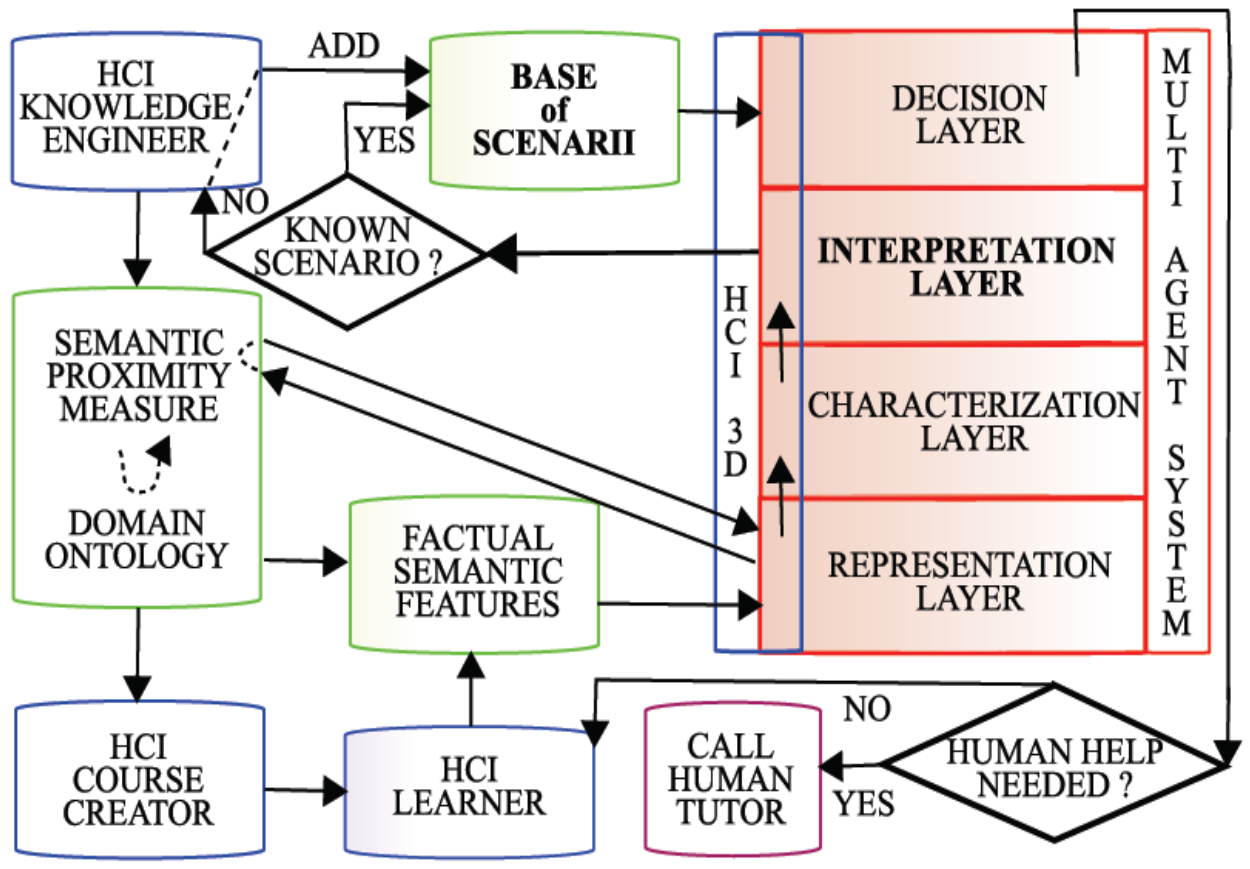

Fig. 2 General architecture of the intelligent tutor

The intelligent tutoring system we propose consists of the three following components (as indicated in Figure 2):

- The graphical interfaces for learners (who are the users for whom the system is developed), for course designers (who must structure the teaching contents) and finally the developers (Human and Computer Interface ‘’HCI’’ knowledge engineer for the knowledge module, and a tree Dimension Human and Computer Interface‘’3D HCI’’ for the behavior of the Multi-Agent Systems);
- The Knowledge module containing: Base of Scenarios, Factual Semantic Features, Semantic Proximity Measure and Domain Ontology;
- The hierarchical MAS with four layers.

Research tasks related to the hierarchical structuring of this MAS were conducted on crisis management [4], emergency logistics [15] and E-learning [10].
The architecture of the intelligent tutor, given in Figure 2, is based on the four components proposed by Wenger [34]:

- The interface with learners;
- The structuring of the domain knowledge in the ontology;
- The modeling of the learner using case based reasoning [38]. This is left to the interpretation layer of the MAS with a supervised learning step of learner’s profile;
- Teaching strategies are associated to the different learners’ profiles. Profiles and teaching strategies are stored in the base of scenarios. The choice of the strategy must be adapted to the situation left to decision layer of the MAS.

The possible recourse to a human tutor is expected. This supposes to detect that the learner is in a situation such as the intervention by human tutor is necessary.

The analysis of the current situation must be carried out in a continuous and dynamic way. Indeed, the treated situation is a layout which evolves over time. The system based on the case based reasoning which we propose, must take this evolution into account. This brings us to the implementation of a system of case based reasoning for dynamic situations. The case based reasoning is the subject of the following section.

## 4. Case-Based Reasoning

Case-Based Reasoning (CBR) is an artificial intelligence methodology which aims at solving new problems based on past experience or the solutions of similar previous problems in the available memory [17]. The solved problems are called source cases and are stored in a database (called a case-base or base of scenarios). The problem to be solved is stored as a new case and is called target case. A CBR is a combination of knowledge and processes to manage and re-use past experience.

The process of Case-Based Reasoning is generally composed of five phases as given in Figure 3: presentation, retrieval, adaptation, validation and update. In the first phase the current problem is identified and completed in such a way that it becomes compatible with the contents and retrieval methods of the case-base. The task of retrieving phase is to find the most similar case(s) to the current problem in the case-base. The goal of the adaptation phase is to modify the solution of case source found in order to build a solution for the target case. The phase of revision, is the step in which the solution suggested in the preceding phase will be evaluated. If the solution is unsatisfactory, then it will be corrected. Finally, the retained step allows to update the knowledge of the system following the reasoning [12], [1].

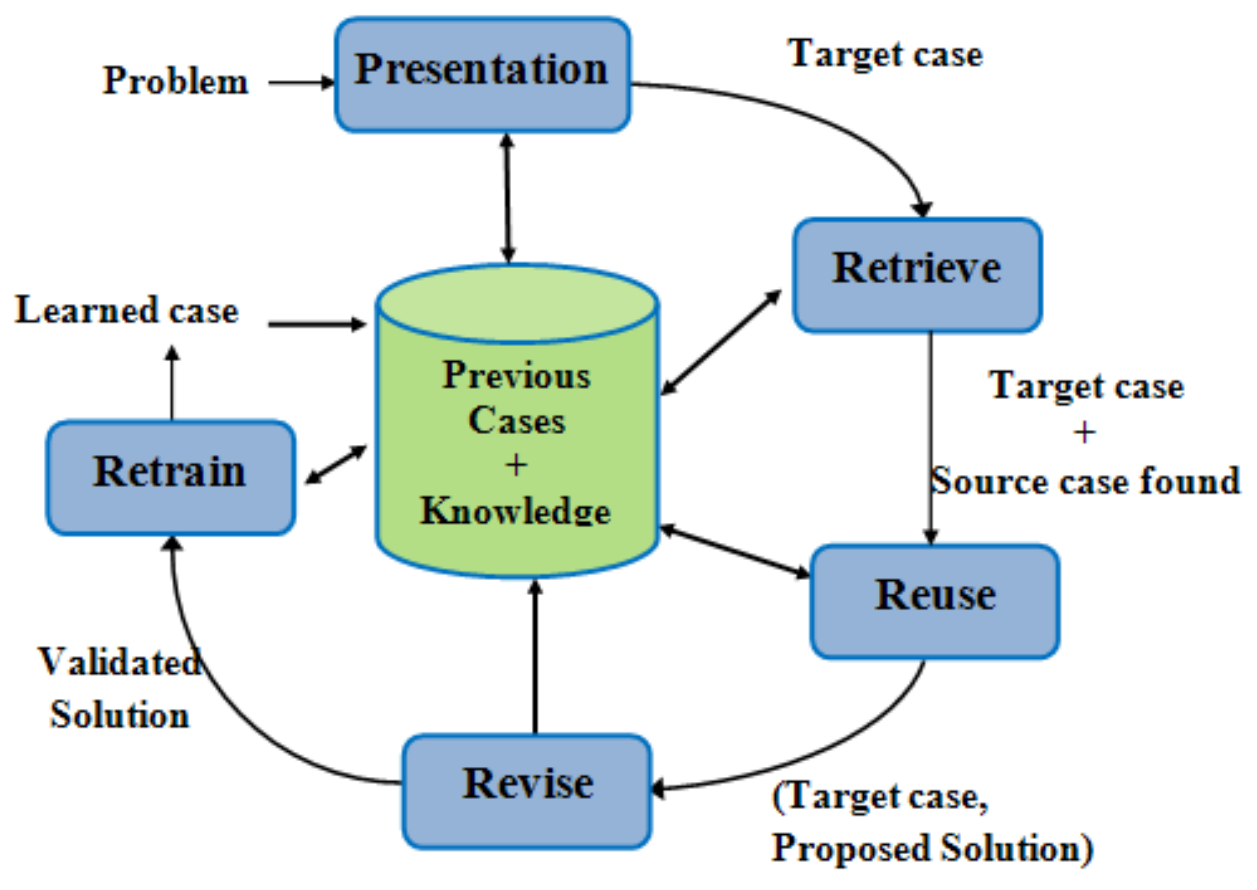

Fig. 3 The CBR cycle (Source [1], [12]).

The systems based on the case-based reasoning can be classified into two categories of applications [19]:

- Applications dealing with situations known as "static". This first model was used with the first CBR systems. Indeed, for this type of system, the CBR static method designer must have all the characteristics describing a case, in advance, in order to be able to realize its model. A data model of the field is thus refined through an expertise in the field of application which can characterize a given situation. Thus, the cases are completely structured in this data model and often represented in a list (a: attributes, v: values) when an attribute is an important specification of the studied field and "v" is the value that is associated with attribute "a" in this case. For example CHIEF [13].

We do not exploit this type of CBR to develop our system. We justify this choice by the fact that in the approach oriented static situation, a problem must be completely described before the search begins in the case base. However in our situation, the traces left by the learner during learning session (the target case) evolve dynamically over time, so we must treat a dynamic situation with some important features.

- Applications with dynamic situations. They differ when we compare them to static cases by the fact that they deal with temporal target cases (the situation), by looking for similar cases (better cases) based on a resemblance between histories (for more details on the subject, the reader may refer to [19])). Several works relate to dynamic case based reasoning such as REBECAS [19].

## 4.1 Multi-Agent Case Based Reasoning

Several architectures case-based reasoning has been applied in Multi-Agent Systems to solve some problems. For example, [14] applies case based reasoning to the predator/prey problem, where each predator can learn cases of the behavior of other agents. Working with the stored case, a predator can predict the movement of other predators so as to enhance their coordination [27].

The Multi-Agent Systems based on case based reasoning are used in many applications areas. They can be classified by several criteria:

- How knowledge is organized within the system (i.e., single vs multiple case bases) [25] ?
- How knowledge is processed by the system (i.e., single vs Multi-Agent execution of the case based reasoning cycle) ?

In the field of Multi-Agent Systems based on case based reasoning, one of the fundamental themes is the autonomy of the agents. Two key factors that govern agent autonomy are (1) its capability to identify whether it is qualified to solve a problem, and (2) its capability to interact with other agents by negotiation and collaboration in order to get a solution for a given problem [25].

In the knowledge processing system, which is the most important criteria, we can distinguish two types of applications:

- The Multi-Agent Systems in which each agent uses the case based reasoning internally to their own needs (level agent case based reasoning) : This type is the first model that was applied in Multi-Agent CBR Systems. For this type of system, each agent is able to find similar cases to the target case in their own case base, also able to accomplish the other steps of CBR cycle. For example we have the system POMAESS in e-service field [36], CCBR framework to personalized route planning [21], and MCBR [18] for distributed systems.

- The Multi-Agent Systems whose approach is a case based reasoning (level Multi-Agent Case Based Reasoning) : For this types of applications, the Multi-Agent Case Based Reasoning System distribute the some/all steps of the CBR cycle (Representation, Retrieve, Reuse, Revise, Retain) among several agents. This type of approach might be better than the first. Indeed the individual agents experience may be limited, therefore their knowledge and predictions too, thus the agents can benefit from the other agents capabilities, cooperate with each other for better prediction of the situation. For example we have the example PROCLAIM [30] in argumentation field, and the Multi-Agent Systems CBR-TEAM [26] approach that uses a set of heterogeneous cooperative agents in a parametric design task (steam-condenser component design).

## 4.2 Multi-Agent Dynamic Case Based Reasoning

Our problem is similar to the CBR for dynamic situations. Indeed, the traces left by the learner during the learning session evolve dynamically over time; the case-based reasoning must take into account this evolution in an incremental way. In other words, we do not consider each evolution of the traces as a new target.
The case-based reasoning which we propose offer important features:

- It is dynamic. Indeed we must continually acquire new knowledge to better reproduce human behavior in each situation.
- It is incremental, this is its major feature because the trace evolves in a dynamic way for the same target case.

The main benefits of our approach are the distributed capabilities of the Multi-Agent Systems and the self-adaption ability to the changes that occur in each situation.

Each action of the learner is represented by a data structure called semantic features that are supported by factual agents. The course of the learner is well represented by a set of trace agents [10]. Therefore, the various actions of the learner (learner traces) can be represented as a collection of semantic features. These will feed the representation layer (Layer 1). The role of this layer is to be both, a picture of the current situation being analyzed and to represent the dynamics of its evolutions over time.

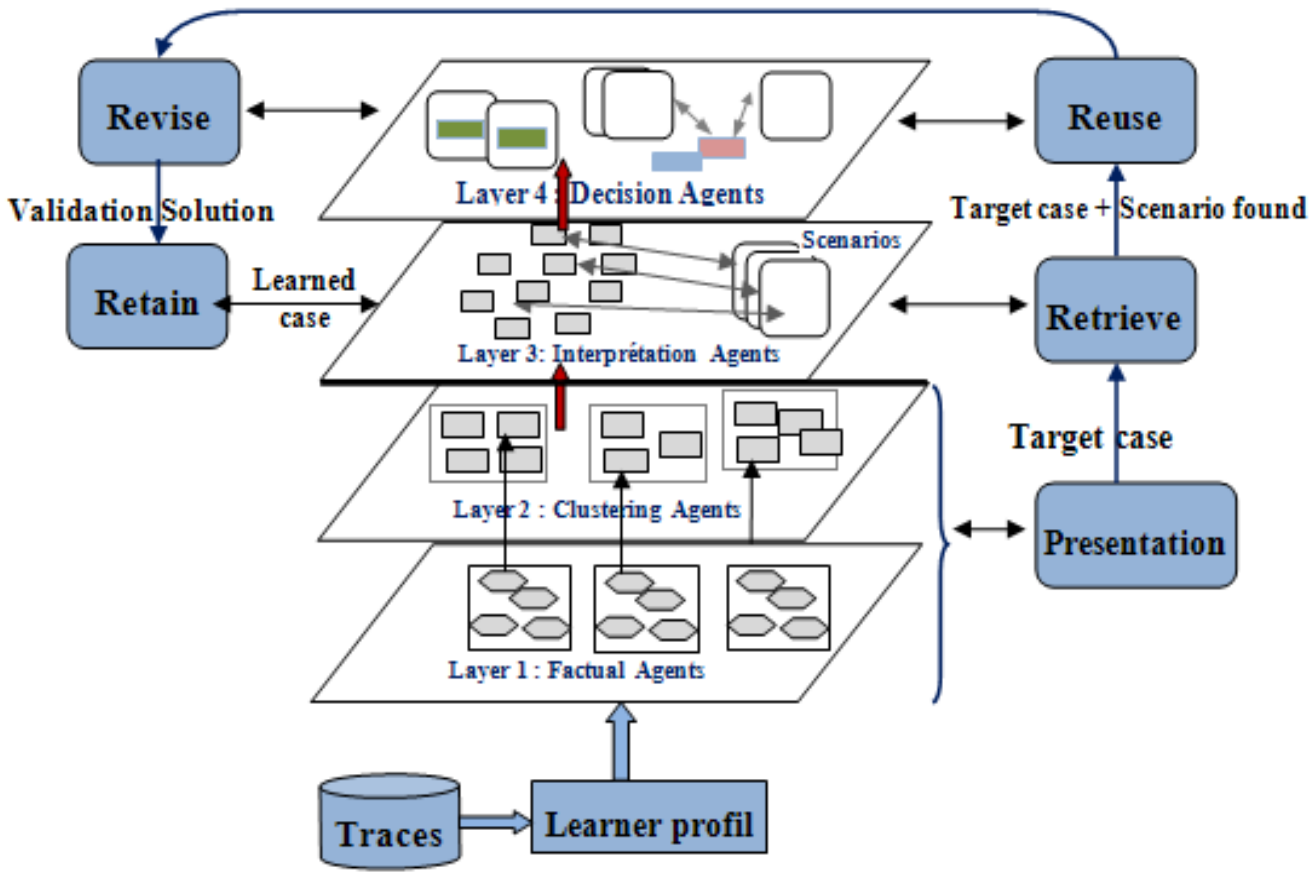

Fig. 4 Dynamic CBR cycle in our approach

The goal of the characterization layer (Layer 2) is to provide a synthetic vision of the organization of agents of the representation layer by classifying them in several subsets according to their activity degrees. A part of the target case in the dynamic and incremental case-based reasoning is developed by this layer.

The interpretation, or prediction, layer (Layer 3) will associate the agents characterization subsets layer with a scenario. The interpretation agents also allow to update the system knowledge by the learning of new cases. In fact, they store and manage new scenarios [10].

The decision layer (Layer 4) selects similar scenarios in the base of scenarios and chooses one to propose to the learner. For each particular situation, the decision agents can react differently depending on the learner's profile concerned, for example, deciding to initiate a communication session with a learner's experiencing difficulties. The human tutor is needed if the system detects a learning situation requiring his/her intervention.

## 4.3 Interpretation Layer

Retrieval of Scenarios is one important step within the case-based reasoning paradigm. The success of retrieval step will depend on three factors: the case representation, case memory and similarity measure used to retrieve scenarios that are similar to the target case (the situation). A several similarity measuring approch have been used in different systems. There is no similarity measure that can accomplish all areas.

There are two ways research for the case in dynamic situations:

- Research by evaluating similarity between the current problem and the already solved problems (the scenarios) in a single dimension [19]. Research in single dimension runs in several stages. Each is used to evaluate the similarity between the current problem and scenarios in a single variable or parameter [2]. Choosing the best case for reuse depends on the results obtained in different steps. Several systems have been used this type of approach such as REBECAS [19] and SAPED [2].

Research by evaluating similarity between the current problem and the already solved problems (the scenarios) in a multiple dimension [2]. The multidimensional research, it is realized in a single step by taking into account all the parameters describing the current problem at the same time. The multidimensional research is also used in several systems, such as CASEP2 [37].

## 4.4 State of the Art on Similarity Measures

Search for similar scenarios are based on the similarity measure. In this part, we present the principles similarity measures often used in case based reasoning, for more details on the subject, the reader my refer to [2].

Biological Sequences Alignment: Dynamic Programming, is an important tool, which has been used for many applications in biology. It is a way of arranging the sequences of DNA, or protein to identify regions of similarity that may be a consequence of structural or functional relationships between the sequences. They are also used in different fields, such as natural language or data mining.

Minkowski distance: The Minkowski distance is a metric on Euclidean space which can be considered as a generalization of both the Euclidean distance.

Longest Common Sub-Sequence (LCSS): the goal is to find the longest subsequence common in two or more sequences [31]. The LCSS is usually defined as: Given two sequences, find the longest subsequence present in both of them. A subsequence is a sequence that appears in the same order, but not necessarily contiguous. The main goal is to count the number of pairs of points considered similar when browsing the two compared sequences.

There are other similarity measures such as Dynamic Time Warping (DTW): The DTW algorithm is able to find the optimal alignment between two sequences. It is often used in speech recognition to determine if two waveforms represent the same spoken phrase. In addition to speech recognition, dynamic time warping has been successfully used in many other fields [16], such as robotics, data mining, and medicine.

Table 1. Comparison of various similarity measures [2]

| | **Type** | **Dimension** | **Length** |
|---|---|---|---|
| **Biological Sequences Alignment** | Symbolic | One-dimensional | Different |
| **DTW** | Digital | One-dimensional | Different |
| **LCSS** | Heterogeneous | Multidimensional | Different |
| **Minkowski distance** | Digital | One-dimensional | Same Length |

## 4.5 Inverse Longest Common Sub-Sequence

The main goal of the retrieval phase in our system is to predict the behavior of the learner, by the reconciliation between the course of the learner (traces in progress or the situation) and past courses (past traces or scenarios). The success of a case-based reasoning system depends primarily on the performance of the retrieval step used and, more particularly, on similarity measure used to retrieve scenarios that are similar to the course of the learner (traces in progress). Several research works have been focused on the similarity measure. Furthermore, these methods are not well suited when we compare two heterogeneous sequences containing textual data (we need semantic distance). In addition we must begin to compare the sequences from tail.

In order to deal with this issue, we propose a complementary similarity measure entitled Inverse Longest Common Sub-Sequence an extension of the Longest Common Sub-Sequence measure [31].

The various actions of the learner (learner traces) can be represented as a collection of semantic features SF=(object, (qualification, value) +), we note object=O, qualification=Q and value=V, SF=(O,(Q,V)+), so the learner traces at time i, can be defined by the formula:

$$LT_i = \bigcup_{1 \le k \le i} SF_k \quad (1)$$

Where $SF_k = (O_k, (Q_{k,1}, V_1), \ldots, (Q_{k,d}, V_d))$ is a sequence of d+1 dimension. Finally the learner traces at time i is a multidimensional sequence.

Let A and B two Traces with size n x d and m x d respectively, where:

$A = ((O_{A,1}, (Q_{A,1,1}, V_{A,1,1}), \ldots, (Q_{A,1,d}, V_{A,1,d}), (O_{A,2}, (Q_{A,2,1}, V_{A,2,1}), \ldots, (Q_{A,2,d}, V_{A,2,d})), \ldots.., (O_{A,n}, (Q_{A,n,1}, V_{A,n,1}), \ldots, (Q_{A,n,d}, V_{A,n,d})))$

and

$B = ((O_{B,1}, (Q_{B,1,1}, V_{B,11}), \ldots, (Q_{B,1,d}, V_{B,1,d}), (O_{B,2}, (Q_{B,2,1}, V_{B,21}), \ldots, (Q_{B,2,d}, V_{B,2,d})), \ldots.., (O_{B,m}, (Q_{B,m,1}, V_{B,m,1}), \ldots, (Q_{B,m,d}, V_{B,m,d}))).$

For a Trace A, let Tail(A) be the Trace:

$Tail(A) = (O_{A,2}, (Q_{A,2,1}, V_{A,2,1}), \ldots, (Q_{A,2,d}, V_{A,2,d})), \ldots.., (O_{A,n}, (Q_{A,n,1}, V_{A,n,1}), \ldots, (Q_{A,n,d}, V_{A,n,d}))).$

Given a real numbers α, β, ε, $\delta$, we define the $ILCSS_{\alpha,\beta,\delta,\varepsilon}(A,B)$ as follows :

$$ILCSS_{\alpha,\beta,\delta,\varepsilon}(A,B) = \begin{cases} 0 \; \textit{if A or B is empty} \\ 1 + ILCSS_{\alpha,\beta,\delta,\varepsilon}(Tail(A), Tail(B)) \text{ if } DS(O_{A,1} - O_{B,1}) \le \alpha \\ DS(Q_{A,1,i}, Q_{B,1,i}) \le \beta,\ |V_{A,1,i} - V_{B,1,i}| \le \beta \text{ for } 1 \le i \le d \\ \max(ILCSS_{\alpha,\beta,\delta,\varepsilon}(Tail(A), B), ILCSS_{\alpha,\beta,\delta,\varepsilon}(A, Tail(B))) \text{ otherwise} \end{cases}$$

Where: $DS(O_{A,1}, O_{B,1})$ is a Symantec distance between the concepts $O_{A,1}$, $O_{B,1}$ and $DS(Q_{A,1,i}, Q_{B,1,i})$ is a Symantec distance between the concepts $Q_{A,1,i}$, $Q_{B,1,i}$ for $1 \le i \le d$.

The CEHL personalization is primarily depending on the ability to produce relevant and exploitable traces of the learner's activity. These traces allow us to describe and to document the learner's activity. They are re-used as a learning support, in order to be able to react during a teaching activity. The learner's traces which feed our system will be the subject of the following section.

## 5. Learner's Traces and Ontology of Course

### 5.1 Learner's Traces

Based on the general definition of a trace given in [25], "a trace is a thing or a succession of things left by an unspecified action and relative to a being or an object; a succession of prints or marks which the passage of a being or an object leaves; it is what one recognizes that something existed; what remains of a past thing". In addition, in CEHL literature, a digital trace is an observed collection, all structured information resulting from an interaction observation temporally located [22].

In our context, a digital trace is resulting from an activity observation representing a process interactional signature. Indeed, it is composed of the objects which are respectively located the ones compared to the others when observed and registered on a support. That means that a trace is explicitly composed of the structured objects and registered compared to a time representation of the traced activity. The structuring can be sequentially explicit (each trace observed is followed and/or preceded by another) or can also come from the temporal characteristic of the traces objects [32]. Indeed, the structuring depends on the type of the time representation and the time of the traced activity. We can distinguish two types of representations:

- They can be a temporal interval determined by two dates, (start and end of observation). In this case, the observed traces activity may be associated with an instant or an interval of time. Then we will be able to take into account chronological relationships between observations';
- They can be a sequence of unspecified elements (for example a sub-part of the whole of the set of integers). In this case, we will focus on the succession or the precedence of the trace observed. Here there is no chronological time.

In the current uses of the traces for the CEHL, collected situations are contrasted: from "we take what we have in well specified formats, what is called the logs" to "we scrupulously instruments the environment to recover the observed controlled and useful for different actors (learner and tutor). The first step consists of modeling the raw data contained in the log file. It is necessary to be able to collect files of traces containing at least, the following elements: time for the start date of the action, codes action which consists in codifying the learner's actions and learner concerned.

Solving the problem of the CEHL personalization is primarily dependent on the capacity to produce relevant and exploitable traces of individual or collective activity of the learner which interacts with a CEHL. For this, we will combine the concepts which can represent all the knowledge of a domain in an explicit and formal specification, by using the domain ontology [10], [39].

### 5.2 Learner's Ontology of Course

Ontology of the Domain: An ontology contains concepts that represent all the knowledge of a domain in an explicit and formal specification [10]. It shows the relationships and rules of associations between these concepts to allow both the system, the production of new knowledge through an inference that the human and system granting of common sense to the terms used in a field of activity to remove any ambiguity during the treatments.

The ontologies become a theme of topical interest within the research conducted in the CEHL. The knowledge diffusion motivation and their acquisitions by learners is central for the CEHL. In this context, the ontologies have a main and indispensable role to take, for sharing and dissemination of the knowledge. The CEHL literature proposes several ontologies for the description of the domain application, the resources and learners. There is thus an resource ontology, an learner ontology and field ontology [10].

Our system needs the knowledge on the learner course to represent it, for this reason, we suggest an ontology of the learner course, able to describe the concepts related to the activities and the traces carried out by the learner at the time of his learning, and recorded on the learning platform: course and its various parts; average and the difficult exercises, lab, the evaluation form, homework, etc.

To build this ontology, we rely on the method developed by [3], which is based on three steps:

- Specify the terms to be collected.
- Organize the terms by using the meta-categories: concepts, attributes, etc.
- Refine ontology and structure it under a hierarchical organization.

The continuous information processing coming from the CEHL allows to suggest to the actors the possible evolutions of the learner work. For that reason, we proceed to the formalization of the information representation received from the environment. To represent the learner activities, it is enough to categorize the various semantic features while being based on ontology.

Semantic Features and Proximity Measures: The semantic feature (SF) is the most basic information which can result from the observation. In other words, the SF cannot be reduced because it is subatomic information and it is structured by respecting an established format. The semantic feature specification allows the viewer to formalize the information communicated to the system.

A SF is a three-part-relation SF= (object, (qualification, value) +) representing a partial aspect of the situation [6]. The SF is composed of the object called selector and its associated qualifiers and their values on the moment of observation. These qualifiers refer to the statements of objects and are incorporated into the ontology of the field. The SF can be enriched in order to situate it in time and space. We can also classify the various SF.

The observations must be grouped, compared, calibrated and differentiated by measuring the similarity and proximity [6]. To bring the same object of a semantic feature observed with two different learners, we must compare the SF on the one hand, by bringing their objects then their qualifiers, and their associated values, on the other hand.

Note that, the proximity is used to evaluate in a quantitative manner the similarity of the objects described by the information resulting from the system observed in the form of semantic features.

The use of semantic features as subatomic granules of information: at a given time, allows to represent the current situation in the form of a collection of semantic features related to the different actions of a learner. These features are the carried by the agents of the representation layer (factual agents) in our system.

## 6. Conclusion and Future Work

Our system allows connecting and comparing the scenario found (current situation) to past scenarios that are stored in a database. The continuous analysis of information coming from the environment (learner's traces) makes it possible to suggest to various actors (learners and tutor) possible evolutions of the current situation.

The Multi-Agent architecture that we propose is based on four layers of agents with a pyramidal relation. The lower layer allows building a representation of the target case, i.e. the current situation. The second layer allows implementing a dynamic and incremental elaboration of the target case. The third layer implements a dynamic process of the source cases recall allowing the search for past situations similar to the current one. Finally, the decision layer captures the responses sent by the interpretation agents to transform them into actions proposed either by machine tutor, virtual tutor, or/and human tutor.

We have presented systems based on Dynamic Case Based Reasoning and we have also clarified that the CBR-based applications can be classified according to the study area: CBR for static situations and CBR for dynamic situations. In our situation, we have used a dynamic case based reasoning with important features. Indeed, the current situation (target case) is a trace that evolves; the case based reasoning must take into account this evolution incrementally. In other words, it shouldn't consider each evolution of the trace as a new target case. In addition, we made a comparison of different existing similarity measures between sequences and we have proposed our new similarity measure (a complementary similarity measure), named Inverse Longest Common Sub-Sequence (ILCSS). Our future work consists in realizing a complete comparative study between our system and other tools.

**Abdelhamid ZOUHAIR** is a PhD student in Cotutelle between the Laboratory LIST, FST of Tangier, Morocco and the Laboratory LITIS, the University of Le Havre, France, since September 2009.

**El Mokhtar EN-NAIMI** is a Professor in Faculty of Sciences and Technologies of Tangier, Department of Computer Science. He is a member of the Laboratory LIST (Laboratoire d'Informatique, Systèmes et Télécommunications), the University of Abdelmalek Essaâdi, FST of Tangier, Morocco. In addition, he is a associate member of the ISCN - Institute of Complex Systems in Normandy, the University of Le Havre, France.

**Benaissa AMAMI** is a Professor in Faculty of Sciences and Technologies of Tangier. He is a Director of the Laboratory LIST (Laboratoire d'Informatique, Systèmes et Télécommunications), the University of Abdelmalek Essaâdi, FST of Tangier, Morocco.

**Hadhoum BOUKACHOUR and Patrick PERSON** are Professors in the University of Le Havre, France. They are members in the Laboratory LITIS (Laboratoire d'Informatique, de Traitement de l'Information et Système), The University of le Havre, France.

**Cyrille BERTELLE** is Professor in the University of Le Havre, France. He is a Deputy Director of the Research Laboratory LITIS at the University of Le Havre and Co-founder of ISCN - Institute of Complex Systems in Normandy, the University of Le Havre, France.